# 支持背景知识的多维端到端短语识别算法研究

李政[1]　涂刚[1]　刘广[1]　占志强[1]　刘译键[1]

[1]（华中科技大学 计算机科学与技术学院　武汉　430074）

（tugang@hust.edu.cn）

## Research on multi-dimensional end-to-end phrase recognition algorithm based on background knowledge

Li Zhen[1], Tu Gang[1], Liu Guang[1], Zhan Zhiqiang[1] and Liu Yijian[1]

[1]（*School of Computer Science and Technology, Huazhong University of Science and Technology*, Wuhan 430074）

**Abstract** At present, the deep end-to-end method based on supervised learning is used in entity recognition and dependency analysis. There are two problems in this method: firstly, background knowledge cannot be introduced; secondly, multi granularity and nested features of natural language cannot be recognized. In order to solve these problems, the annotation rules based on phrase window are proposed, and the corresponding multi-dimensional end-to-end phrase recognition algorithm is designed. This annotation rule divides sentences into seven types of nested phrases, and indicates the dependency between phrases. The algorithm can not only introduce background knowledge, recognize all kinds of nested phrases in sentences, but also recognize the dependency between phrases. The experimental results show that the annotation rule is easy to use and has no ambiguity; the matching algorithm is more consistent with the multi granularity and diversity characteristics of syntax than the traditional end-to-end algorithm. The experiment on CPWD dataset, by introducing background knowledge, the new algorithm improves the accuracy of the end-to-end method by more than one point. The corresponding method was applied to the CCL 2018 competition and won the first place in the task of Chinese humor type recognition.

摘要　目前，实体识别与依存关系分析，采用的主要是基于监督学习的深度端到端方法。这种方法存在两个问题：首先，不能引入背景知识；其次，不能识别出自然语言的多粒度、嵌套特征。为了解决这些问题，提出了基于短语窗口的标注规则，同时设计了配套的多维端到端短语识别算法。该标注规则以短语为最小单位，把句子分成 7 类可嵌套的短语类型，同时标示出短语之间的依存关系。配套设计的算法，不仅可以引入背景知识，识别出句子中的各类嵌套短语片段，而且可以识别出短语之间的依存关系。实验的结果表明，该标注规则方便易用，没有二义性；配套算法比传统端到端算法更加符合语法的多粒度与多样性特征，在 CPWD 数据集上实验，引入背景知识，比端到端方法结合词性维度准确性提高 1 个点以上。对应方法应用到 CCL2018 比赛中，在中文幽默类型识别任务中取得第一名。

关键词　自然语言处理；标注体系；短语识别；依存分析

## 1. 引言

随着即时通信、微博、论坛、朋友圈等的快速流行，人们在网络上发布的文字信息也越来越多。这些文字信息蕴含极大的价值，它们是分析社会整体及公众个体的观点、喜好、情绪、趋势等的入口。快速准确地分析文字信息，是自然语言处理研究的目标。

目前，实体识别与依存关系分析，采用的主要是基于监督学习的深度端到端方法。这种方法存在一些不足。首先，不能借助背景知识，比如："中华人民共和国"是知识库中已知实体名称，但是算法模型不能方便的导入这些知识库信息；其次，不能很好的识别出语言的多粒度、嵌套特征。比如："中华人民共和国国务院"，包括"中华人民共和国"、"国务院"、"中华人民共和国国务院"等多粒度、嵌套的实体名称，但是端到端模型只能预测出单一的标签；再者，一个端到端模型只能完成一类预测任务，不能同时预测命名实体以及他们之间的依存关系。

为了解决这些问题，只能从源头上入手：改变标注规则，同时设计更加合适的算法。本文首先提出了基于短语窗口的标注规则。该标注规则既可以表示语言的多粒度、嵌套关系，又可以表示语言片段之间的依存关系。然后，使用该规则，标注了各种类型的句子数据集，我们把这个数据集称为中文短语窗口数据集（Chinese Phrase Window Dataset，CPWD）。最后，设计了相应的算法。新的算法采用多维输入和多维输出的端到端模型，可以把背景知识作为信息输入，同时，识别出多粒度、嵌套短语，以及短语之间的依存关系，对应模型称为多维端到端模型（Multi Dimensional End-to-End Model，MDM）。实验的结果表明，该标注规则方便易用，没有二义性；MDM 模型比端到端模型更加适用于语法的多粒度与多样性特征，准确性有明显提高。

## 2. 相关工作

语块分析体系最早是由 Abney 在 1991 年提出的语块描述体系[1]，之后 Kudoh 等[2]提出了一种基于支持向量机的语块自动分析方法；同时，Shen 等[3]提出了一种投票分类策略，将多种不同的数据表示和多种训练模型结合在一起，根据投票分类策略确定最终结果；此外，Mancev 等[4]提出了一种处理支持向量机非凸结构的斜率损失的最小化问题的序列双向方法。在汉语的语块分析方面，周强等[5]构造了基于规则的汉语基本块分析器，并设计了相应的基本块规则，给出了一整套解决方案，提高了基于规则的基本块分析器的性能；此外，李超等[7]应用最大熵模型和马尔科夫模型构建了一套汉语基本块的分布识别系统。

深度学习方法出现后，短语识别研究迎来了快速发展。Chiu 等[8]使用双向 LSTM 提取文本全局特征，同时，使用 CNN 提取单词的特征，进行名词短语实体的识别；Kuru 等[9]使用 Stacked Bidirectional LSTMs 提取文本全局特征进行名词短语实体识别，取得了较大进展；侯潇琪等[10]利用深度模型，将词的分布表征作为模型的输入特征维度，用于基本短语识别任务中，比使用传统的词特征表示方法提高明显；李国臣等[11]以字作为标注单元和输入特征，基于深层模型研究短语的识别问题，并将基于 C&W 和 Word2Vec 两种方法训练得到的字分布表征作为模型的特征参数，避免了对分词及词性标注结果的依赖；徐菁等[12]利用知识图谱，提出基于主题模型和语义分析的无监督的名词短语实体指称识别方法，同时具备短语边界检测和短语分类功能；程钟慧等[13]提出了一种基于强化学习的协同训练框架，在少量标注数据的情况下，无须人工参与，利用大量无标注数据自动提升模型性能，从非结构化大数据集中抽取有意义的名词短语。

语法依存最早是著名的法国语言学家特思尼耶尔在《怎样建立一种句法》一书谈到，我国学者徐烈炯等[14]认为，语义角色是一个"句法-语义"接口概念，而不是单纯的语义概念；刘宇红[15]提出语义和语法双向互动的观点；孙道功[16]基于词汇义征和范畴义征的分析，研究了词汇与句法的衔接机制；亢世勇等[17]通过构建"现代汉语句法语义信息语料库"，研究了义类不同的体词在施事（主语、宾语、状语）和受事（主语、宾语、状语）六个语块的分布特点。这其中还包括哈工大、百度、清华等团队的语法分类贡献。

在语法分析方面，McDonald 等[18]提出了基于图模型的依存句法分析器 MSTParser；Nivre 等[19]提出了基于转移模型的依存句法分析器 MaltParser；Ren 等[20]对 MaltParser 依存句法分析器的 Nivre 算法进行了优化，有效的改进了在汉语中难以解决的长距离依存等问题；车万翔等[21]对 MSTParser 依存句法分析器进行了改进，使用了图模型中的高阶特征，提高了依存句法分析的精度；Chris 等[22]在基于转移模型的依存句法分析框架上运用长短时记忆神经网络，将传统的栈、队列、转移动作序列看作 3 个 LSTM 细胞单元，将所有转移的历史均记录在 LSTM 中，改进了长距离依存问题；Tao Ji 等[23]开发了一种依赖树节点表示形式，可以捕获高阶信息，通过使用图神经网络（GNN），解析器可以在 PTB 上实现最佳的 UAS 和 LAS；Yuxuan Wang 等[24]提出了一种基于神经过渡的解析器，通过使用基于列表的弧跃迁过渡算法的一种变体，进行依赖图解析，获得了较好的效果；Fried 等[25]通过强化学习来训练基于过渡的解析器，提出了将策略梯度训练应用于几个选区解析器的实验，包括基于 RNN 过渡的解析器。

在语义分析方面，丁伟伟等[24]利用 CRF 在英文语料上能够利用论元之间的相互关系、提高标注准确率的特点，将其运用到中文命题库，使用 CRF 对中文语义组块分类，取得好的效果；王丽杰等[25]提出了基于图的自动汉语语义分析方法，使用哈工大构建

的汉语语义依存树库完成了依存弧和语义关系的分析；王倩等[26]基于谓词和句义类型块，使用支持向量机的语义角色对句子的句义类型进行识别，也有一定的启发意义。

综上，传统方法存在一些不足。首先，不能引入背景知识；其次，不能很好识别出语言的多粒度、嵌套特征；再者，端到端模型只能完成一类预测任务，不能同时预测命名实体以及他们之间的依存关系。

为了解决这些问题，我们从源头上入手：改变标注规则，同时设计更加合适的算法。本文首先提出了基于短语窗口的标注规则。然后，使用该规则，标注了各种类型的句子数据集 CPWD。最后，设计了相应的多维端到端模型 MDM。

## 3. 短语标注规范

为了实现句子的短语识别与依存关系分析，制定了一套短语标注规范。该短语标注规范不仅可以对多粒度、嵌套短语进行标注，而且可以反映短语之间的依存关系。标注规则相对简单，容易推广。

根据该规则，我们标注了中文短语窗口数据集 CPWD，数据集包括 45,000 条从对话、新闻、法律、政策、小说中挑选的非文言文的中文句子。为了方便模型设计，句子最大长度限制在 50 个字以内。

### 3.1. 短语依存关系

标注规范将句子中的短语分成：名词短语、动词短语、数量词短语、介词短语、连词短语、语气词、从句，总共 7 类基本类型。句子由短语组成，因此 7 类短语类型通过树状结构组成句子，即依存关系。

通常，句子的树状结构由主、谓、宾关系组成，图 1 是句子语法树结构图。a)句子成分树：句子"我爱祖国"，按照句子语法可以分为主语"我"、谓语"爱"、宾语"祖国"；b)句子原型树：把"我"、"爱"、"祖国"放到对应的主谓宾位置；c)短语类别树："我"是名词短语，"爱"是动词短语，"祖国"是名词短语。

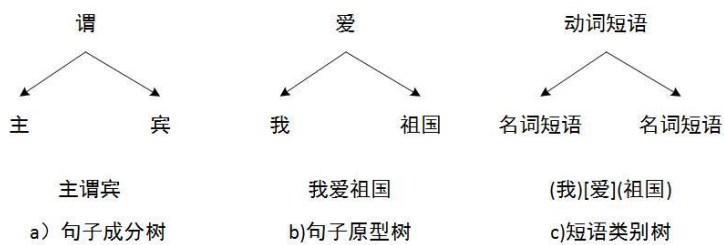

Fig.1 Syntax tree structure
图 1. 句子语法树结构图

对于复杂的句子同样可以采用这种方法进行短语识别和依存关系的分析。图 2 是复杂句子的语义单元划分过程。为了方便介绍，我们使用"()"表示名词短语，"[]"表示动词短语，"{}"表示数量词短语，"<>"表示介词短语，"##"表示连词短语，"@@"表示语气词短语，"\^"表示从句。

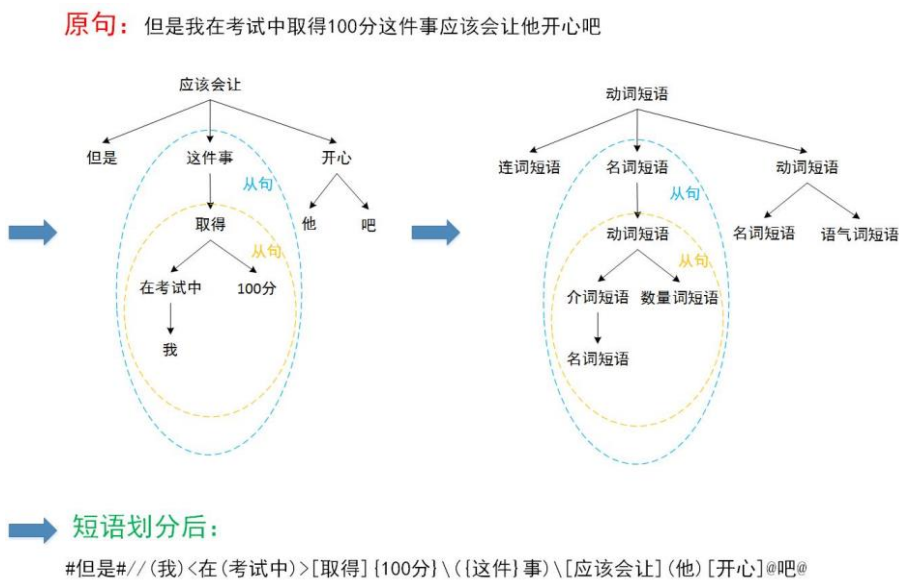

Fig.1 Phrase dependency
图 2. 短语依存关系

## 3.2. 短语标注规范

短语类别有名词短语"()"、动词短语"[]"、数量词短语"{}"、介词短语"<>"、连词短语"##"、语气词短语"@@"、从句"/\"。标注允许嵌套结构的存在，每种短语类别的标注规则如下。

(1) 连词短语：

连词短语是用来连接词与词、词组与词组或句子与句子、表示某种逻辑关系的虚词。连词短语可以表并列、承接、转折、因果、选择、假设、比较、让步等关系。

如："但是"表转折，"因为""所以"表因果等。

在标注体系中连词短语一般无嵌套关系。

(2) 语气词短语：

语气词短语是表示语气的虚词，常用在句尾或句中停顿处表示种种语气。

如："吗"、"吧"、"呢"等在词语、句子末，表示语气。

在标注体系中语气词短语一般无嵌套关系。

(3) 名词短语：

表示人或事物以及时间、方位等，在句子中主要充当主语、宾语、定语。它包括：以名词为中心词的偏正短语（如："伟大祖国"，"这些孩子"）；用名词构成的联合短语（如："工人农民"）；复指短语（如："首都北京"）；方位短语（如："桌面上"，"大楼前面"）；"的"字短语（如："打更的老汉"）等。某些名词短语的中心词也可以是动词、形容词，定语可以是代词、名词或其他名词短语。

(4) 动词短语：

动词短语代表动作，包括起修饰作用的状语与补语。

如："马上开始了"包括状语"马上"，中心动词"开始"，补语"了"。

(5) 介词短语：

又称为介宾短语，是介词和其它代词或名词或从句搭配形成的短语。

- "在这次考试中"为介词+名词短语，标注为<在({这次}考试中)>。
- "被"、"把"字句。如：<被(他)>，<把(他)>。

(6) 数量词短语：

数量短语，指由数词和量词组合构成的短语。

- 数量词和名词搭配，如：({一首}动听的曲子)
- 作为状语，如：[{一蹦一跳}地走着]
- 作为补语，如：[看了](他){一眼}

(7) 从句：

为了标注一个完整的语义单元，需要使用从句结构体现短语间的层次关系。

- 兼语句标为从句。如："我命令他去外面"，这里"他"既是前面的宾语，又是后面的主语，标注：(我)[命令]/(他)[去](外面)\。
- 连动句标为从句。如："我出去骑车打球"，"骑车"、"打球"为连动，标注为：(我)//[出去]\/[骑](车)\/[打](球)\\。
- 主语从句、宾语从句。如：(他)[说]/(计算机)[正在改变](世界)\。

## 4. 算法

本节详细介绍短语多维端到端算法。短语多维端到端算法采用多维输入，多维输出的形式。我们以句子长度 50 为例，例句为"我的祖国是中华人民共和国"。

多维输入包括 50 维，其中，第 n 维对应的是知识库中的长度为 n 的短语的信息，例如：长度为 7 的"中华人民共和国"是知识库中的实体，第 7 维输入对应于"中"字的偏移位置，会放入实体分类的标记。标记类型包括名词、动词、介词等多种词性；

多维输出包括 50 维，其中，第 n 维对应预测结果中的长度为 n 的短语类型（7 类），例如：长度为 7 的"中华人民共和国"被预测成了名词，则第 7 维输出对应于"中"字的偏移位置开始的 7 个标签，会预测为名词标记。标记类型包括 7 种短语类型。

可见，算法可以预测多粒度、嵌套短语，以及短语依存关系。下面几节详细介绍算法对应的模型 MDM，包括算法模型基本结构图，以及具体的背景知识、分类网络、短语分片及依存关系的应用。

### 4.1. 算法流程

算法流程如下。

(1) 数据标注。例如：原句为"我爱祖国"，标注后为"(我)[爱](祖国)"。

(2) 输入维度。使用"我爱祖国"字向量作为输入。字向量以外的 50 维，需要引入知识库查找结果：第 1 维对应标签包括："我"标记为"名词"，"爱"标记为"动词"；第 2 维对应标签包

括：" 祖国"标记为"名词"；如果有没找到的字词，则标记为 0。这里并不关注分词正确性，只要是知识库找到的词，都会作为输入，即使是错误的分词，由模型训练判断对错。字词查找采用 AC（Aho-Corasick automaton）树算法。

(3) 短语特征提取。通过特征提取网络进行特征提取，可以使用语言模型进行特征提取。

(4) 短语分类。使用全连接识别出短语类别与依存关系。例如："我"是名词短语，"爱"动词短语，"祖国"名词短语。

(5) 输出维度。将 50 维短语分类预测结果综合，去除有冲突的预测，比如长度与维度不匹配错误等，使用标注符号表示出来，得到结果。例如："(我)[爱](祖国)"。

### 4.2. MDM 模型

短语算法模型采用多维输入与多维输出的端到端模型，识别短语类别及他们之间的依存关系。图 3 是模型结构图。主要分为四个部分：（1）字向量和背景知识层：字向量是将句子中的字进行嵌入的一种方式；背景知识是使用知识库查询结果进行标示；字向量与背景知识 concat 后通过 embedding 层输入；（2）特征提取层：特征提取层负责特征的抽取，该层使用 Bi-LSTM 或者 BERT 语言模型；（3）分类层：使用特征提取层提取的特征进行全连接分类，实现短语类别预测；（4）短语分片及依存关系层：对句子中的短语进行识别，同时解决冲突，预测出依存关系。

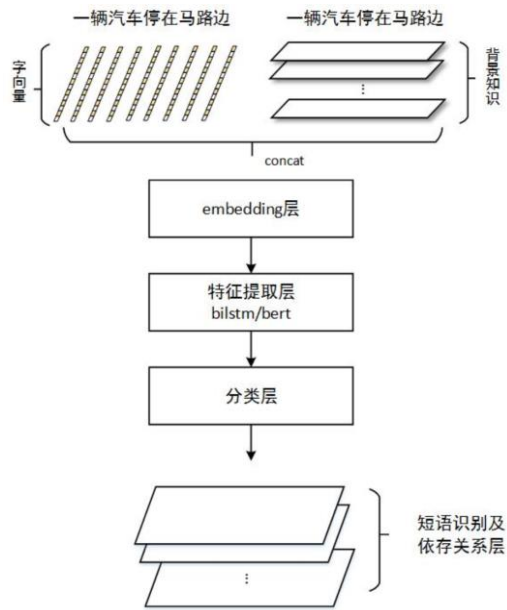

Fig.3 Algorithm structure
图 3. 算法结构

(1) 字向量和背景知识嵌入层

模型输入是字向量和背景知识。字向量是通过 word2vec 训练得到的一组向量，每个字向量是 256 维组成。

背景知识以标记形式输入，包括分词与词性信息。标记包含词性信息，并且，标记以 B 标示开始位置，而且只在开始位置标示，与分词的 BIO 方式不同，见表 1。

Table.1 Input tag type
表 1 输入标记类型

| 名词 | 动词 | 形容词 | 数词 | 量词 | 介词 | 连词 | 语气 | 拟声 | 助词 |
|---|---|---|---|---|---|---|---|---|---|
| B-ming | B-dong | B-xing | B-shu | B-liang | B-jie | B-lian | B-yu | B-ni | B-zhu |

背景知识是增加的维度，由 50 维的知识编码形成。其中，第 n 维对应的是知识库中的长度为 n 的短语的信息，例如：句子"一辆汽车停在马路边"中，长度为 2 的词包括："一辆""汽车""马路""路边"，长度为 3 的词："马路边"，这些是从知识库中的查找获得的。那么，第 2 维输入对应于词的第一个字"一""汽""马""路"的位置，会放入实体分类的标记，标记类型包括名词、动词、

介词等多种词性（这里为了方便图示，简化成统一用 B 代替）。第 3 维输入对应"马"的位置，放入实体分类的标记，代表"马路边"是一个长度为 3 的词。

背景知识引入了一句话的不同组成成分在知识库中的信息。增加背景知识后模型能学到更多的语言特征，可以引入更多的信息来进行监督学习，提高短语识别和依存关系的准确性。引入的知识库分类信息并不需要关注分词结果，而是查找到了对应的词就把输入信息加入相应维度，由模型训练判断取舍。如："马路边"在该句里面的正确分词是"马路"而不是"路边"，但是由于知识库中存在"路边"一词，所以还是引入到输入的维度信息中，让模型训练去习得规律。

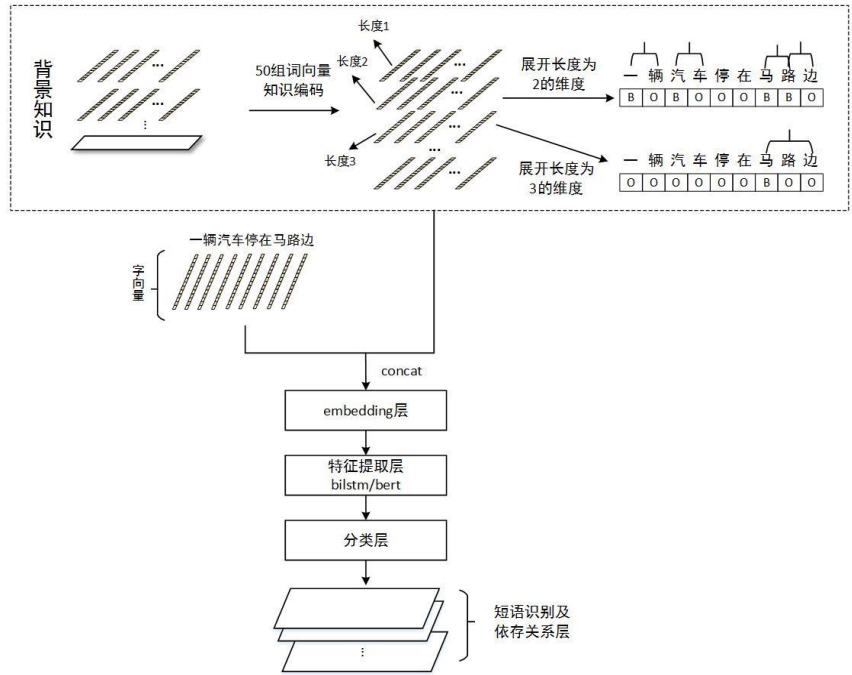

Fig.4 Background knowledge

图 4. 背景知识

实验证明，这种处理方式可以省略分词步骤，而且结果比分词后进行预测更准确。

由于模型输入的句子长度限定为 50，所以有 50 维的输入知识编码，对应从最短的（长度为 1）的短语到最长的（长度为 50）的短语。如图 5 所示，第 n 维背景知识表示的是长度为 n 的词的知识编码，其中第一维、第二维背景知识大多是一些名词、动词、介词等基本词分类。第三维以后，随着长度变长，短语大多是一些实体名称，经过训练，发现模型对于实体名称的权重设置较大，遇到长于 3 的实体名称，都可以正确识别，而且可以识别出多粒度、嵌套的实体名称与各类短语。

用 B-ming、B-dong 等标示名词、动词的起始位置，如图 5 的方式进行输入编码："马路边"是名词，"在马路边"是介词短语，"一辆"是数词，"一辆汽车"是名词。嵌套与依存关系都蕴含在这种标示中。

| 输入: | 维度 | 一 | 辆 | 汽 | 车 | 停 | 在 | 马 | 路 | 边 |
|---|---|---|---|---|---|---|---|---|---|---|
| | | | | | | 标记 | | | | |
| | 1 | 0 | 0 | 0 | 0 | B-dong | B-jie | 0 | 0 | 0 |
| | 2 | B-shu | 0 | B-ming | 0 | 0 | 0 | B-ming | B-ming | 0 |
| | 3 | 0 | 0 | 0 | 0 | 0 | 0 | B-ming | 0 | 0 |
| | 4 | B-ming | 0 | 0 | 0 | 0 | B-jie | 0 | 0 | 0 |
| | 5 | 0 | 0 | 0 | 0 | 0 | 0 | 0 | 0 | 0 |
| | 6 | 0 | 0 | 0 | 0 | 0 | 0 | 0 | 0 | 0 |
| | ...... | | | | | | | | | |

Fig.5 Examples of multi-dimensional background knowledge input tags

图 5. 多维背景知识输入标记举例

(2) 特征提取层

特征提取层负责短语特征的提取，特征提取比较成熟，可以选用卷积神经网络、双向长短记忆网络 Bi-LSTM、BERT 等来实现特征抽取。目

前测试最佳的网络架构是 BERT 与 Bi-LSTM 的组合。经过特征提取网络后，输出的是 512 维的隐层权值向量，特征提取提取的结果将被用于后续的分类网络。

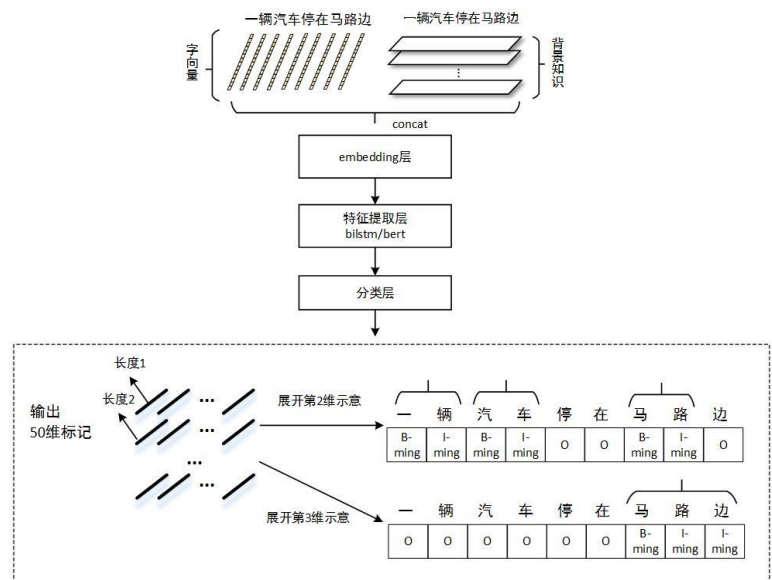

Fig.6 Phrases and dependency recognition

图 6. 短语及依存关系识别

(3) 分类层

对特征提取层提取到的特征进行全连接，输出 50 维的预测标签，输出第 n 维对应的是长度为 n 的短语；标签是分词常用的 B/I 组合方式，其中 B 代表开始字，I 代表后续的字，如：三个字"马路边"在第三维输出，对应的标签是"B-ming I-ming I-ming"，这里与输入层只用 B 作为标记不同，如图 6。

多维输出包括 50 维，其中第 n 维对应预测结果中的长度为 n 的短语类型（7 类），例如：长度为 7 的"中华人民共和国"被预测成了名词，则第 7 维输出对应于"中"字的偏移位置，会输出名词标记，标记类型包括 7 种短语类型，如表 2。

Table.2 Output tag type

表 2 输出标记类型

| 名词短语 | 动词短语 | 数量词短语 |
|---|---|---|
| B/I-ming | B/I-dong | B/I-shu |
| 介词短语 | 连词 | 语气 |
| B/I-jie | B/I-lian | B/I-yu |

(4) 短语识别及依存关系层

该层对判断后的短语类型进行输出。

输出 50 维的短语识别和依存关系结果，其中有错误的地方。比如：第 2 维输出了一个"B-ming I-ming I-ming"长度为三个字的预测标签，即使采用 crf 层也无法避免，需要把这种标签去掉。此外，还有预测冲突与错误的标签，比如：以 I 开头的标签。该层对这些问题进行统一整理，然后输出一致的预测标签，同时生成类似"(我)[爱](祖国)"的预测结果。

(5) 损失函数

损失函数，由输出的所有维度决定。采用交叉熵函数，通过交叉熵来计算预测的结果和标签之间的距离。式 1 是损失函数，其中 $M$ 是短语类别数量，$y_c$ 是真实分布，$p_c$ 是预测结果的分布。总的 loss 是 50 维 loss 的求和。

$$loss = -\sum_{c=1}^{M} y_c \log(p_c) \quad (1)$$

5. 实验

5.1. 数据集与评估标准

实验数据集使用标注的中文短语窗口数据集 CPWD，包括 45,000 条从对话、新闻、法律、政策、小说中挑选的中文句子。其中文言文只占不到 5%比例，多是一些成语与谚语组成的句子。为了方便模型设计，句子最大长度限制在 50 个字以内。

传统端到端的结果统计，主要根据每个字预测得到标签的情况。这样的统计方式存在偏差，不如按照短语统计准确，比如："中华人民共和国"命名实体的标签是"BIIIIII"为正确，如果预测结果是"BIIIBII"，那么存在一个标签错误。按照传统方式统计，只算 7 个标签中出现了一次错误；按照短语方式统计，

"BIIIIII"全对为正确，"BIIIBII"为错误，即有一个标签错误，整个短语的预测是错误的，这样更加准确合理。

## 5.2. 实验环境

实验采用 Python 语言实现，python 版本为 3.6.1。使用的框架为 TensorFlow，版本为 1.12.0。使用的电脑配置为内存：32G，处理器：Intel Xeon(R) CPU E5-2623 v3 @3.00GHz*8，显卡：TITAN Xp，操作系统类型：ubuntu14.04 64-bit。

## 5.3. 实验分析

首先，MDM 在多种网络结构或者数据形式的情况下进行对比优化，同时，在准确性和时间上做一个均衡。优化方法包括采用双层 BiLSTM，采用 BERT 代替 BiLSTM，使用 CRF 层，在 BiLSTM 层之前加入 CNN 层进行特征抽取，在特征抽取层之后加入 Transformer 结构，选取不同比例的反例，输入维度不同等。结果如表 3 所示。

Table.3 MDM optimization process comparision
表 3. MDM 优化过程对比

|  | Recall | F1 | 前向传播时间(ms) |
|---|---|---|---|
| 1 层 BiLSTM | 87.47 | 86.27 | **65** |
| 2 层 BiLSTM | 88.70 | 87.12 | 80 |
| BERT | 89.06 | 88.60 | 106 |
| BiLSTM+CRF | 87.33 | 87.62 | 101 |
| BERT+CRF | **89.31** | **89.95** | 124 |
| CNN+BiLSTM | 86.42 | 86.56 | 96 |
| BiLSTM+Trans | 88.68 | 89.10 | 93 |
| 1:1 正反例 | 87.52 | 87.35 | 86 |
| 1:2 正反例 | 88.14 | 88.66 | 88 |
| 5 维输入 | 87.31 | 88.30 | 68 |
| 50 维输入 | 88.14 | 88.54 | 93 |

从结果可以看出，最优的模型结构是：50 维输入，1:2 正反例，BERT+CRF。但是考虑到运行效率和资源占用情况，我们在做实验或者工程部署的时候，建议采用的模型是：50 维输入，1:2 正反例，BiLSTM+Transformer。

Loss 值收敛情况在 BERT、BiLSTM 两种最优模型之间进行比较，收敛情况如图7,可以看到,BiLSTM 收敛快些，BERT 语言模型的收敛慢，一个 epoch 的训练时间也长些。在实际部署中，我们推荐使用 BiLSTM 作为特征提取层，可以提高效率。

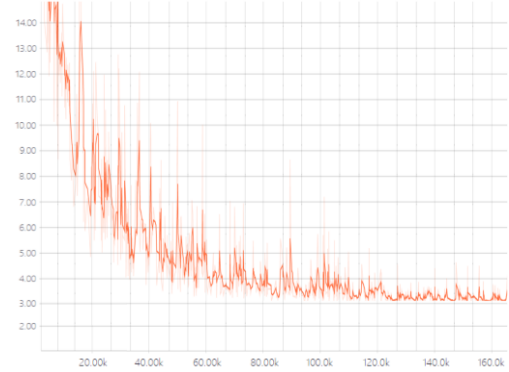

（a）BERT

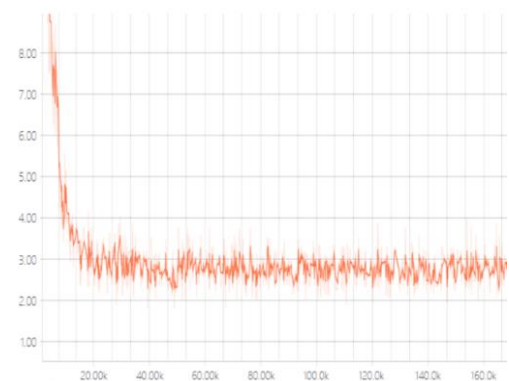

（b）BiLSTM

Fig.7 Loss Convergence
图 7. Loss 收敛情况图

## 5.4. 对比实验结果

MDM 对比各种端到端算法。

由于各种端到端算法输出与标签形式有不同，所以将标签形式调整成在统一的方式下进行对比。MDM 模型输出标签比端到端多,附带有嵌套等信息，所以需要进行降维处理，之后与端到端模型进行对比。降维后可以形成命名实体标签、依存标签，分别与 BiLSTM、BERT 等端到端算法进行对比。

Table.4 Comparative experiment result
表 4. 对比实验结果

|  | 命名实体 F1 | 依存分析 F1 |
|---|---|---|
| BiLSTM | 88.12 | 86.70 |
| BiLSTM+CRF | 89.06 | 87.99 |
| BERT | 90.82 | 88.20 |
| BERT+CRF | 90.32 | 89.10 |
| CNN+CRF | 88.41 | 87.89 |
| MDM（BiLSTM） | 91.25 | 89.15 |
| MDM（BERT） | **92.60** | **91.67** |

从表 4 可以看到 MDM 模型比传统端到端有优

势。这种优势的产生,我们分析是由于 MDM 更适合语言多样性,使得模型不用在嵌套的命名实体之间做多选一的抉择,降低了模型的困惑度;同时,MDM 可以预测语法依存关系,模型不需要进行降维映射,所以就不会造成特征的丢失,也因此获得了更高的准确性。

## 6. 总结

针对传统端到端算法的一些问题,我们从源头上入手:改变标注规则,同时设计更加合适的算法。首先提出了基于短语窗口的标注规则。该标注规则既可以标示语言的多粒度、嵌套关系,又可以标示语言片段之间的依存关系。然后,使用该规则,标注了包括各种类型句子的数据集 CPWD。最后,设计了多维端到端模型,可以把背景知识作为信息输入,同时,识别出多粒度、嵌套短语,以及短语之间的依存关系。实验的结果表明,该标注规则方便易用,没有二义性;MDM 模型比端到端模型更加适用于语法的多粒度与多样性特征,引入背景知识后,准确性有明显提高。

## 参考文献